\documentclass[letterpaper]{article} 
\usepackage[draft]{aaai2027}  
\usepackage[hyphens]{url}  
\usepackage{graphicx} 
\usepackage{natbib}  
\usepackage{caption} 
\usepackage{comment}

\usepackage{amsmath,amssymb,amsfonts}
\usepackage{booktabs}
\usepackage{multirow}
\usepackage{makecell}

\usepackage{xcolor}

\title{CRS-Triage: Confidence- and Reliability-Aware Selective Triage under Incomplete Clinical Evidence}

\author{
    Guan Qiang\textsuperscript{\rm 1},
    Yushen Chen\textsuperscript{\rm 1},
    Tianlong Liu\textsuperscript{\rm 1,*},
    David Rotenberg\textsuperscript{\rm 2},
    Ethan H. Kim\textsuperscript{\rm 2},
    Fang Fang\textsuperscript{\rm 1,*}
}

\affiliations{
    \textsuperscript{\rm 1}Western University,
    London, Ontario, Canada\\
    \textsuperscript{\rm 2}Centre for Addiction and Mental Health,
    Toronto, Ontario, Canada\\
    gqiang@uwo.ca, yche2692@uwo.ca,
    tianlong.liu@uwo.ca,\\
    David.Rotenberg@camh.ca, Ethan.Kim@camh.ca,
    fang.fang@uwo.ca\\
}

\begin{document}

\maketitle

\begingroup
\renewcommand{\thefootnote}{*}
\footnotetext{Corresponding authors}
\endgroup

\begin{abstract}
Emergency triage requires reliable decisions within a short time period. However, the available electronic health record (EHR) data, including structured data and clinical text, are often incomplete, unreliable, and inconsistent. This makes machine learning (ML)-based triage prediction more challenging, as existing ML models typically rely on complete and reliable EHR data to accurately predict acuity levels of patients. To address this, we propose confidence- and reliability-aware selective triage (CRS-Triage) to predict patients’ acuity levels with a confidence score. By comparing the confidence score with a predefined threshold, the proposed CRS-Triage can selectively determine whether the model should make the decision or defer the case. Specifically, CRS-Triage separately evaluates the reliability of structured data and clinical text and then jointly considers the consistency between the
two modalities to estimate the confidence of each prediction. Moreover, to reduce the risk of missing high-acuity patients, i.e., under-triage, the proposed CRS-Triage prefers to assign patients slightly higher acuity levels, i.e., over-triage, by penalizing under-triage errors. Experiments on the MIMIC-IV-ED dataset show that CRS-Triage achieves strong prediction performance. It also provides a better
risk--coverage trade-off and remains reliable when the available
EHR data is incomplete, degraded, or inconsistent across
modalities.
\end{abstract}

\section{Introduction}
Emergency triage requires determining the acuity level of each patient encounter quickly and accurately. To achieve this, reliable clinical information is needed. However, in practice, the available electronic health record (EHR) data is often incomplete, unreliable, or even inconsistent across different data sources ~\citep{xie2022benchmarking,soenksen2022integrated}. Specifically, in the emergency department, the EHR data for each encounter typically consist of structured data and short text data. However, structured data, such as vital signs and intake information, may be missing and unreliable at the time of triage. Short text fields, such as chief complaints and triage notes, may also vary in length and quality~\citep{mitra2023learning}. For instance, some notes contain clear clinical information, while others are brief, unclear, or even inconsistent with the structured data. 

Although ML-based triage prediction methods are widely used, they may
still produce unreliable predictions because they do not explicitly
estimate the reliability of each modality or jointly model predictive
uncertainty.
Existing ML-based methods handle the two modalities of EHR data separately. In structured data, missing values are often handled through imputation or masking before prediction~\citep{lee2023learning}. For clinical text, it is typically cleaned and tokenized and then represented with methods such as TF-IDF or pretrained language models. Subsequently, the processed structured and text data are commonly combined through concatenation, late fusion, or attention-based fusion~\citep{thao2024medfuse,yao2024drfuse}. However, these fusion methods do not explicitly consider the reliability of each modality. As a result, unreliable information may still have a substantial influence on the fused representation that leads to inaccurate or overconfident predictions. Moreover, in existing ML-based triage prediction methods, uncertainty is typically estimated only after the prediction model has been trained, rather than being jointly learned with it~\citep{tomani2022parameterized}. Thus, the estimated uncertainty may not reliably indicate how likely
a prediction is to be incorrect or whether accepting it could lead to
a clinically unsafe decision.

Another challenge in existing research is that under-triage can delay urgently needed care and therefore poses a greater risk to patient safety. In triage, there are typically two types of errors, including under-triage and over-triage. Under-triage occurs when a high-acuity patient is assigned a lower acuity level, which may delay treatment. In contrast, over-triage happens when a low-acuity patient is assigned a higher acuity level, which may consume additional clinical resources but is generally less harmful~\citep{huabbangyang2023associated}. Therefore, while a triage model should aim to classify patients' acuity levels as accurately as possible, it should treat these two types of errors differently. When classification errors cannot be completely avoided, the model should favor over-triage over under-triage to reduce the risk of delaying necessary care for high-acuity patients. Existing methods balance these two errors through class weighting, cost-sensitive loss functions, or decision-threshold adjustment to place greater emphasis on identifying high-acuity patients and reducing under-triage~\citep{wyatt2024leveraging,lin2024interpretable}. However, the associated weights, loss parameters, or decision thresholds are typically fixed at the population level and cannot adapt to the quality of EHR data or predictive uncertainty on a case-by-case basis.

To address the above-mentioned challenges, we propose CRS-Triage, a confidence- and reliability-aware selective model for emergency triage. CRS-Triage calculates the reliability of the structured data and clinical text for every patient encounter. It then evaluates the consistency between the two modalities by comparing their prediction distributions for each modality. The estimated modality reliability and cross-modal consistency are then used to determine the weights of each modality during fusion, and the fused prediction distribution is used to predict the patient’s acuity level. To reflect the embedded uncertainty for each prediction, we propose a confidence score that is calculated using modality reliability and cross-modal consistency. Finally, the confidence score is compared with a threshold to determine whether the model should issue a prediction or defer the case for further assessment. Overall, our proposed CRS-Triage remains reliable under incomplete and unreliable EHR data by making accurate predictions based on an assessment of the reliability of the available EHR data. The main contributions are summarized as follows:
\begin{itemize}
    \item A reliability-aware multimodal fusion mechanism is developed to estimate the reliability of structured data and clinical text for each patient encounter. The estimated reliability of each modality is jointly considered with cross-modal consistency during fusion, therefore reducing the influence of unreliable information on the final prediction.

    \item A confidence score calculation mechanism is utilized to determine whether to accept a model-based prediction or defer the case for further assessment. Therefore, the risk of overconfident predictions can be reduced.

    \item  Larger penalties for under-triage errors are designed to prevent the model from assigning lower acuity levels to high-acuity patients. Thus, the risk of delaying urgent care for high-acuity patients is reduced.
\end{itemize}

\section{Related Work}

\subsection{Emergency Triage with Incomplete Data}

ML-based models have been widely used in emergency triage
to predict patient acuity from structured EHR data,
including demographics and vital signs
~\citep{levin2018machine,xie2022benchmarking,
lin2024interpretable}.
While most studies focus on predictive performance,
recent work has also examined triage errors across
patient groups~\citep{sax2023evaluation,wyatt2024leveraging}.
However, the reliability of structured evidence is
rarely assessed at the encounter level.
Missingness is common in EHRs and may reflect clinical
measurement processes~\citep{che2018recurrent,groenwold2020informative}.
Existing approaches address missing data through
imputation, missingness masks, and models for incomplete
time series, while recent work has further examined
its relationship with patient outcomes~\citep{liang2025causal}.
Although these methods improve the handling of incomplete
records, they do not explicitly quantify the reliability
of the remaining structured evidence for individual
predictions. Consequently, predictions from incomplete
records may be treated as equally reliable as those
from more complete records.

\subsection{Clinical Text and Multimodal EHR Learning}
Clinical text complements structured EHR data by capturing
information not fully represented by clinical variables.
Existing studies combine these modalities through feature
concatenation, attention, or transformer-based fusion
~\citep{soenksen2022integrated,thao2024medfuse}.
However, multimodal EHR data are often incomplete.
Recent methods learn modality-aware representations from
incomplete records~\citep{lee2023learning}.
M3Care estimates missing information from available
modalities~\citep{zhang2022m3care}, while MUSE and FlexCare
support varying modality and label availability
~\citep{xu2024flexcare}.
DrFuse further learns shared and modality-specific
representations to address missing modalities
~\citep{yao2024drfuse}.
Although effective, these methods primarily address
missing modalities or joint representation learning,
rather than differences in reliability when both
modalities are available.
In emergency triage, structured features may be missing,
clinical text may be brief or ambiguous, and the two
modalities may provide conflicting predictions.
Evidential multi-view methods estimate predictive
evidence and uncertainty while handling cross-modal
disagreement~\citep{han2021trusted,xu2024reliable},
but do not explicitly quantify modality-specific
reliability.

\subsection{Uncertainty and Selective Prediction}
Beyond modality reliability, confidence in fused predictions
must also be trustworthy. Even accurate models may produce
overconfident predictions under incomplete or shifted data
~\citep{guo2017calibration}.
Evidential learning models class probabilities using a
Dirichlet distribution and estimates uncertainty from
predictive evidence~\citep{sensoy2018evidential}.
However, evidential uncertainty may not reliably identify
incorrect predictions~\citep{shen2024uncertainty}.
Selective prediction allows uncertain cases to be deferred
~\citep{geifman2019selectivenet}, including to human experts
~\citep{mozannar2020consistent},
whose decisions may depend on how deferral is communicated
~\citep{bondi2022role}.

Emergency triage also involves asymmetric clinical costs:
under-triage may delay urgent care, whereas over-triage
primarily increases resource use
~\citep{huabbangyang2023associated,sax2023evaluation}.
Recent work has addressed this asymmetry through cost-aware
selective prediction using calibrated probabilities and
conformal prediction~\citep{kwon2026conformal}.
However, this approach relies on a single prediction source,
with uncertainty estimation and deferral performed after
model training rather than jointly learned with prediction.

\section{Method}

\begin{figure*}[t]
    \centering
    \includegraphics[
        width=1\textwidth,
        height=0.24\textheight
    ]{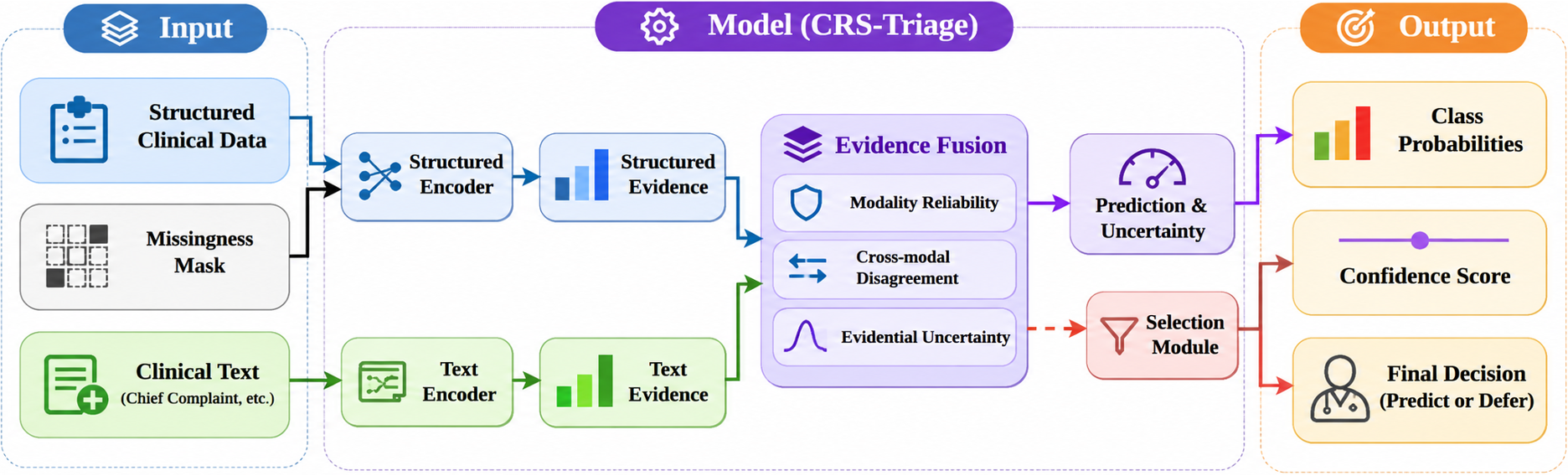}
    \caption{CRS-Triage: reliability- and
    uncertainty-aware multimodal evidence fusion
    for selective emergency triage.}
    \label{fig:crs_triage_overview}
\end{figure*}

\subsection{Problem Setup}

We consider a dataset $\mathcal{D}
=
\{(\mathbf{x}_i^s,\mathbf{m}_i,x_i^t,y_i)\}_{i=1}^{N}
$ with $N$ emergency triage encounters. Let $r\in\{s,t\}$ denote the structured or
text modality. For each encounter $i$, $\mathbf{x}_i^s \in \mathbb{R}^d$ denotes the structured clinical data with a number of $d$ features, which may contain missing values. Let $\mathbf{m}_i \in \{0,1\}^d$ represent the corresponding missingness mask with each entry $m_{i,j}$ to be $1$ if the $j$-th structured feature is observed and $0$ otherwise. The clinical text is denoted by $x_i^t$, and $y_i\in\{1,\ldots,C\}$ denotes the true triage class. The triage classes are ordered in descending order of clinical acuity,
such that lower class indices indicate higher acuity levels. For each encounter, a CRS-Triage model $F_w(\cdot)$, parameterized by $w$, produces a class probability vector $\mathbf{p}_i$
and a confidence score $s_i\in[0,1]$. Specifically, $p_{i,c}$ is the predicted probability that encounter $i$ belongs to triage class $c$ and $\hat{y}_i$ denotes the predicted acuity level.
Given a confidence threshold $\tau\in[0,1]$, the final model output can be represented as:
\begin{equation}
    \tilde{y}_i
=
\begin{cases}
\hat{y}_i, & s_i\geq\tau,\\
\mathrm{defer}, & s_i<\tau.
\end{cases}
\end{equation}

In addition to prediction accuracy, the patient safety risk posed by under-triage should also be considered. Given the adopted label ordering, prediction errors can be divided into under-triage and over-triage. Specifically, $\hat{y}_i>y_i$ indicates under-triage, whereas $\hat{y}_i<y_i$ indicates over-triage. Since under-triage may delay urgently needed care, it is assigned a larger penalty than over-triage. Accordingly, a penalty function for triage errors is defined as:
\begin{equation}
A(y_i,c)
=
\begin{cases}
\lambda_{\mathrm{under}}|c-y_i|, & c>y_i,\\
\lambda_{\mathrm{over}}|c-y_i|, & c<y_i,\\
0, & c=y_i,
\end{cases}
\end{equation}
where $y_i$ is the true class, $c$ is a possible predicted class, and
$\lambda_{\mathrm{under}}
>
\lambda_{\mathrm{over}}
>
0.$
Thus, the expected penalty for triage errors under the predictive distribution is expressed as $
\mathcal{C}_i
=
\sum_{c=1}^{C}A(y_i,c)p_{ic}$.

\subsection{Modality-Specific Encoding and Prediction}

After imputation, the structured feature vector is denoted by
$\tilde{\mathbf{x}}_i^s\in\mathbb{R}^d$. Let $\phi_v(\cdot)$ and
$\phi_m(\cdot)$ denote the learnable projection functions for the
imputed values $\tilde{\mathbf{x}}_i^s$ and the missingness mask
$\mathbf{m}_i$, respectively. The two inputs are jointly encoded as $
\mathbf{z}_i^s
=
W_sE_s
\left(
[
\phi_v(\tilde{\mathbf{x}}_i^s);
\phi_m(\mathbf{m}_i)
]
\right)
+b_s$. Here $E_s(\cdot)$ is the structured-data encoder, $W_s$ and $b_s$ are the learnable parameters, and $\left[\cdot,\cdot\right]$ denotes the concatenation operation. For the text modality, the clinical text $x_i^t$ is tokenized by
$\operatorname{Tok}(\cdot)$ and encoded by the pretrained text encoder
$E_t(\cdot)$. The resulting token representations are aggregated using the mean-pooling function $\operatorname{MeanPool}(\cdot)$ with the token-level attention mask
$\boldsymbol{\mu}_i^t$~\citep{reimers2019sentencebert}. Similarly, the learnable
projection matrix $W_t$ and bias vector $b_t$ are then used to obtain
the fixed-length text representation $\mathbf{z}_i^t$: $
\mathbf{z}_i^t
=
W_t
\operatorname{MeanPool}
\left(
E_t(\operatorname{Tok}(x_i^t));
\boldsymbol{\mu}_i^t
\right)
+
b_t$. After the two modalities have been encoded, each modality produces its
own predictive evidence. The structured-data prediction head $G_s(\cdot)$ and text prediction
head $G_t(\cdot)$ map the representations $\mathbf{z}_i^s$ and
$\mathbf{z}_i^t$ to the evidence vectors $\mathbf{e}_i^s$ and
$\mathbf{e}_i^t$, respectively. A softplus activation is applied to
each output to ensure that both evidence vectors are nonnegative
~\citep{xu2024reliable}.
These evidence vectors
are then used to construct the modality-specific predictive
distributions.
Following Dirichlet-based evidential learning
~\citep{sensoy2018evidential}, the evidence
$\mathbf{e}_i^r$ from each modality $r\in\{s,t\}$ is converted into
the Dirichlet parameters
$\boldsymbol{\alpha}_i^r=\mathbf{e}_i^r+\mathbf{1}$, where
$\mathbf{1}\in\mathbb{R}^C$ is a vector of ones. The total Dirichlet
strength is calculated as
$S_i^r=\sum_{c=1}^{C}\alpha_{ic}^r$. The modality-specific class
probability and evidential uncertainty are then given by
$p_{ic}^r=\alpha_{ic}^r/S_i^r$ and $u_i^r=C/S_i^r$, respectively.
Finally, the corresponding probability vector is denoted by $
\mathbf{p}_i^r
=
(p_{i1}^r,\ldots,p_{iC}^r)$.

\subsection{Reliability-Aware Evidential Fusion}
After the modality-specific predictions and uncertainty estimates are obtained, CRS-Triage estimates the reliability of each modality before fusion. For the structured data, reliability is calculated from feature availability. Since structured features may contribute differently to triage
prediction, an unconstrained learnable parameter $\eta_j$ is introduced
for each feature $j\in\{1,\ldots,d\}$. This parameter is converted into
the nonnegative feature weight
$a_j=\operatorname{softplus}(\eta_j)$. The same feature weights are
used for all encounters. Then, the structured-data reliability score $\rho_i^s$ that represents the weighted proportion of available features, can be calculated as:
\begin{equation}
\rho_i^s
=
\frac{\sum_{j=1}^{d}a_jm_{i,j}}
{\sum_{j=1}^{d}a_j}.
\end{equation}

For the clinical text, reliability is calculated using text length, prediction certainty, and evidential certainty. The normalized text
length $L_i\in[0,1]$ is calculated from the number of non-padding
tokens. The entropy of the text-based prediction can be calculated by  
$\mathcal{H}_i^t=-\sum_{c=1}^{C}p_{ic}^t\log p_{ic}^t$. Based on this
entropy, the prediction certainty is represented by
$1-\mathcal{H}_i^t/\log C$. Then, the evidential certainty can be
represented by $1-u_i^t$. The text reliability score $\rho_i^t$ can be calculated by combining
these three quantities using the trainable weight vector
$\mathbf{w}_{\rho}^{t}$ and bias $b_{\rho}^{t}$:
\begin{equation}
\rho_i^t
=
\sigma
\left(
(\mathbf{w}_{\rho}^{t})^\top
\left[
L_i,\,
1-\frac{\mathcal{H}_i^t}{\log C},\,
1-u_i^t
\right]
+
b_{\rho}^{t}
\right),
\end{equation}
where $\sigma(\cdot)$ is the sigmoid function and both $\rho_i^s$ and
$\rho_i^t$ belong to $[0,1]$.

However, even when the two modalities are reliable individually, they may still
provide conflicting predictions. CRS-Triage therefore evaluates their
cross-modal consistency by comparing the structured-data probability
vector $\mathbf{p}_i^s$ and text-based probability vector
$\mathbf{p}_i^t$~\citep{xu2024reliable}. Specifically, the cross-modal
disagreement score $\delta_i$ is calculated using the normalized
Jensen--Shannon divergence:
\begin{equation}
\delta_i
=
\frac{
\operatorname{JS}
\left(
\mathbf{p}_i^s
\Vert
\mathbf{p}_i^t
\right)
}{
\log 2
}.
\end{equation}
Based on modality reliability, evidential certainty, and cross-modal
consistency, the structured-data weight $q_i^s$ and text weight
$q_i^t$ are calculated by combining these quantities using the
trainable weight vectors $\mathbf{w}_q^s$ and $\mathbf{w}_q^t$ and
biases $b_q^s$ and $b_q^t$, respectively:
\begin{equation}
\begin{aligned}
q_i^s
&=
\sigma
\left(
(\mathbf{w}_{q}^{s})^\top
\left[
\rho_i^s,\,
1-u_i^s,\,
1-\delta_i
\right]
+
b_q^s
\right),\\
q_i^t
&=
\sigma
\left(
(\mathbf{w}_{q}^{t})^\top
\left[
\rho_i^t,\,
1-u_i^t,\,
1-\delta_i
\right]
+
b_q^t
\right).
\end{aligned}
\end{equation}
Unlike normalized attention weights, $q_i^s$ and
$q_i^t$ are estimated independently and are not required to sum to
one. Therefore, both modalities can contribute little evidence when
their information is unreliable, uncertain, or conflicting
~\citep{liu2022trusted}. The estimated modality weights are then applied to the structured-data
evidence $\mathbf{e}_i^s$ and text evidence $\mathbf{e}_i^t$. The
resulting fused evidence vector $\mathbf{e}_i$ can be calculated as $
\mathbf{e}_i
=
q_i^s\mathbf{e}_i^s
+
q_i^t\mathbf{e}_i^t$.

Following the same Dirichlet transformation used for the
modality-specific evidence, the fused evidence $\mathbf{e}_i$ is
converted into the final Dirichlet parameters
$\boldsymbol{\alpha}_i=\mathbf{e}_i+\mathbf{1}$. The corresponding
total Dirichlet strength is
$S_i=\sum_{c=1}^{C}\alpha_{ic}$. The final class probability and fused
uncertainty are then calculated as $p_{ic}=\alpha_{ic}/S_i$ and
$u_i=C/S_i$, respectively.

\subsection{Confidence-Based Selective Prediction}

After the final class probability vector $\mathbf{p}_i$ and fused
uncertainty $u_i$ are obtained, CRS-Triage calculates a confidence
score for each prediction. To measure how clearly the two most likely
classes are separated, let $p_i^{(1)}$ and $p_i^{(2)}$ denote the
largest and second-largest entries of $\mathbf{p}_i$, respectively.
The prediction margin is then calculated as
$\Delta_i=p_i^{(1)}-p_i^{(2)}$. The confidence score $s_i$ can be calculated with the trainable
weight vector $\mathbf{w}_{\mathrm{conf}}$ and bias
$b_{\mathrm{conf}}$:
\begin{equation}
\begin{aligned}
s_i
=
\sigma\Big(
\mathbf{w}_{\mathrm{conf}}^\top
\big[
&\rho_i^s,\,
\rho_i^t,\,
1-u_i^s,\,
1-u_i^t,\\
&1-u_i,\,
1-\delta_i,\,
\Delta_i
\big]
+
b_{\mathrm{conf}}
\Big).
\end{aligned}
\end{equation}
During inference, the
confidence score $s_i$ is compared with a confidence threshold
$\tau\in(0,1)$. The model issues a prediction when $s_i\geq\tau$ and
defers the case for further assessment when $s_i<\tau$.

To make the confidence score reflect the patient safety risk posed by under-triage, the expected triage penalty
$\mathcal{C}_i$ defined above is included during training. For our dataset with $N$ encounters, a target coverage
$\kappa\in(0,1]$ is used to prevent the model from deferring too many
cases. The penalty coefficient $\beta>0$ controls the coverage
constraint, while the small constant $\epsilon>0$ provides numerical
stability. Thus, the selective training loss $\mathcal{L}_{\mathrm{sel}}$ is
defined as:
\begin{equation}
\mathcal{L}_{\mathrm{sel}}
=
\frac{
\sum_{i=1}^{N}s_i\mathcal{C}_i
}{
\sum_{i=1}^{N}s_i+\epsilon
}
+
\beta
\left[
\max
\left(
0,\,
\kappa-\frac{1}{N}\sum_{i=1}^{N}s_i
\right)
\right]^2.
\end{equation}
Here, to prevent the model from assigning low confidence to all cases,
we introduce a coverage constraint that applies a penalty when the
average confidence falls below the target coverage $\kappa$.

\subsection{Training Objective}

All components of CRS-Triage are trained jointly using the evidential
learning, triage penalty, confidence-based selective, and cross-modal
disagreement objectives defined above. Following evidential learning
~\citep{sensoy2018evidential,chen2024think}, a KL-regularized
evidential loss is used for classification. For a Dirichlet parameter vector $\boldsymbol{\alpha}$ and its
one-hot class label $\mathbf{y}$, the total Dirichlet strength can be calculated as
$S=\sum_{c=1}^{C}\alpha_c$. Then, the modified Dirichlet parameter vector can be
defined as
$\tilde{\boldsymbol{\alpha}}
=\mathbf{y}+(\mathbf{1}-\mathbf{y})\odot\boldsymbol{\alpha}$, where
$\mathbf{1}$ is a vector of ones and $\odot$ denotes element-wise
multiplication. The digamma function is denoted by $\psi(\cdot)$, while
$\operatorname{Dir}(\cdot)$ and $\operatorname{KL}(\cdot\Vert\cdot)$
denote the Dirichlet distribution and KL divergence, respectively. The
nonnegative coefficient $\lambda_{\mathrm{KL}}$ controls the KL
regularization. Thus, the per-sample evidential loss
$\ell_{\mathrm{EDL}}(\boldsymbol{\alpha},\mathbf{y})$ is defined as:
\begin{equation}
\begin{aligned}
\ell_{\mathrm{EDL}}(\boldsymbol{\alpha},\mathbf{y})
={}&
\sum_{c=1}^{C}
y_c
\left[
\psi(S)-\psi(\alpha_c)
\right] \\
&+
\lambda_{\mathrm{KL}}
\operatorname{KL}
\left[
\operatorname{Dir}(\tilde{\boldsymbol{\alpha}})
\Vert
\operatorname{Dir}(\mathbf{1})
\right].
\end{aligned}
\end{equation}

For our training set containing $N$ encounters, the evidential loss is
calculated using the fused Dirichlet parameters
$\boldsymbol{\alpha}_i$ and the modality-specific Dirichlet parameters
$\boldsymbol{\alpha}_i^s$ and $\boldsymbol{\alpha}_i^t$. The
nonnegative coefficient $\lambda_{\mathrm{aux}}$ controls the
contribution of the modality-specific supervision. The total
evidential loss $\mathcal{L}_{\mathrm{evi}}$ is calculated as:
\begin{equation}
\begin{aligned}
\mathcal{L}_{\mathrm{evi}}
&=
\frac{1}{N}\sum_{i=1}^{N}
\Bigg[
\ell_{\mathrm{EDL}}
(\boldsymbol{\alpha}_i,\mathbf{y}_i)\\
&\quad+
\frac{\lambda_{\mathrm{aux}}}{2}
\Big(
\ell_{\mathrm{EDL}}
(\boldsymbol{\alpha}_i^s,\mathbf{y}_i)
+
\ell_{\mathrm{EDL}}
(\boldsymbol{\alpha}_i^t,\mathbf{y}_i)
\Big)
\Bigg].
\end{aligned}
\end{equation}

The expected triage penalty $\mathcal{C}_i$ defined above is also
minimized over all training encounters. The corresponding average
triage penalty is
$\mathcal{L}_{\mathrm{pen}}
=N^{-1}\sum_{i=1}^{N}\mathcal{C}_i$. This term reduces the expected
triage penalty over all encounters. The fused uncertainty should also reflect disagreement between the
structured data and clinical text. Therefore, the disagreement loss
$\mathcal{L}_{\mathrm{dis}}$ is introduced to increase the fused
uncertainty $u_i$ when the cross-modal disagreement $\delta_i$ is
large. The positive coefficient $\gamma$ controls the required
relationship between disagreement and uncertainty:
\begin{equation}
\mathcal{L}_{\mathrm{dis}}
=
\frac{1}{N}
\sum_{i=1}^{N}
\left[
\max
\left(
0,\,
\gamma\delta_i-u_i
\right)
\right]^2.
\end{equation}
This loss encourages the fused uncertainty $u_i$ to increase as the
disagreement between the two modalities becomes stronger. Finally, the overall
training objective $\mathcal{L}$ is defined as:
\begin{equation}
\mathcal{L}
=
\mathcal{L}_{\mathrm{evi}}
+
\lambda_{\mathrm{pen}}\mathcal{L}_{\mathrm{pen}}
+
\lambda_{\mathrm{sel}}\mathcal{L}_{\mathrm{sel}}
+
\lambda_{\mathrm{dis}}\mathcal{L}_{\mathrm{dis}}.
\end{equation}

\section{Experiments and Results}

\begin{table*}[!t]
\centering
\begingroup
\small
\setlength{\tabcolsep}{3pt}

\begin{tabular}{@{}llccccc@{}}
\toprule
Dataset & Method
& Macro-F1 $\uparrow$
& QWK $\uparrow$
& UT (\%) $\downarrow$
& Penalty $\downarrow$
& ECE $\downarrow$ \\
\midrule

\multirow{8}{*}{MIMIC-IV-ED}
& Structured only
& $0.604\pm0.006$ & $0.735\pm0.004$
& $19.2\pm0.4$ & $0.698\pm0.008$ & $0.067\pm0.002$ \\

& Text only
& $0.618\pm0.011$ & $0.746\pm0.006$
& $18.5\pm0.6$ & $0.693\pm0.007$ & $0.061\pm0.003$ \\

& Gated fusion
& $0.688\pm0.006$ & $0.801\pm0.004$
& $14.5\pm0.3$ & $0.546\pm0.005$ & $0.041\pm0.002$ \\

& DrFuse
& $0.697\pm0.005$ & $0.812\pm0.004$
& $13.9\pm0.3$ & $0.523\pm0.005$ & $0.038\pm0.002$ \\

& TMC
& $0.704\pm0.006$ & $0.819\pm0.006$
& $13.4\pm0.4$ & $0.509\pm0.005$ & $0.034\pm0.001$ \\

& Base evidential fusion
& $0.707\pm0.005$ & $0.822\pm0.005$
& $13.1\pm0.3$ & $0.499\pm0.007$ & $0.031\pm0.002$ \\

& Unified EHR Transformer
& $0.715\pm0.008$ & $0.828\pm0.006$
& $12.6\pm0.4$ & $0.482\pm0.009$ & $0.037\pm0.003$ \\

& \textbf{CRS-Triage}
& $\mathbf{0.742\pm0.006}$ & $\mathbf{0.846\pm0.003}$
& $\mathbf{9.6\pm0.1}$ & $\mathbf{0.406\pm0.004}$
& $\mathbf{0.021\pm0.001}$ \\

\midrule

\multirow{8}{*}{MC-MED}
& Structured only
& $0.581\pm0.009$ & $0.708\pm0.007$
& $20.8\pm0.6$ & $0.754\pm0.011$ & $0.073\pm0.004$ \\

& Text only
& $0.592\pm0.012$ & $0.721\pm0.009$
& $20.1\pm0.7$ & $0.741\pm0.013$ & $0.069\pm0.004$ \\

& Gated fusion
& $0.663\pm0.008$ & $0.782\pm0.006$
& $16.1\pm0.4$ & $0.606\pm0.008$ & $0.049\pm0.003$ \\

& DrFuse
& $0.674\pm0.007$ & $0.795\pm0.006$
& $15.4\pm0.4$ & $0.581\pm0.008$ & $0.045\pm0.002$ \\

& TMC
& $0.682\pm0.008$ & $0.802\pm0.007$
& $14.9\pm0.5$ & $0.566\pm0.009$ & $0.041\pm0.002$ \\

& Base evidential fusion
& $0.686\pm0.007$ & $0.807\pm0.006$
& $14.5\pm0.4$ & $0.551\pm0.008$ & $0.038\pm0.002$ \\

& Unified EHR Transformer
& $0.695\pm0.009$ & $0.815\pm0.007$
& $13.8\pm0.5$ & $0.533\pm0.011$ & $0.044\pm0.003$ \\

& \textbf{CRS-Triage}
& $\mathbf{0.718\pm0.008}$ & $\mathbf{0.832\pm0.005}$
& $\mathbf{11.2\pm0.3}$ & $\mathbf{0.458\pm0.007}$
& $\mathbf{0.027\pm0.002}$ \\

\bottomrule
\end{tabular}

\endgroup
\caption{Overall performance on MIMIC-IV-ED and MC-MED
at full coverage.}
\label{tab:overall}
\end{table*}

\subsection{Experimental Setup}
CRS-Triage was evaluated on two emergency department datasets,
MIMIC-IV-ED v2.2~\citep{johnson2023mimic} and
MC-MED~\citep{kansal2025mcmed}.
MIMIC-IV-ED contains $418{,}100$ adult encounters from
$201{,}252$ patients with five-level acuity labels.
MC-MED provides an independent cohort with the same
five-level triage labels.
Both datasets use demographics, arrival mode, and triage
vital signs as structured inputs and chief complaints
as text inputs. Pain scores were also included for
MIMIC-IV-ED when available.
Post-triage variables were excluded to prevent data leakage.
Each dataset was independently split into training,
validation, and test sets at the patient level
(70\%/10\%/20\%) to avoid patient overlap.

A two-layer neural network encoded structured data,
while BioClinicalBERT~\citep{alsentzer2019publicly}
encoded clinical text.
Baselines included structured-only and text-only models,
gated fusion, DrFuse~\citep{yao2024drfuse},
trusted multi-view classification (TMC)~\citep{9767662},
base evidential fusion, and a unified EHR
Transformer~\citep{lyu2022multimodal}.
SelectiveNet~\citep{geifman2019selectivenet} and a
cost-aware selection method were also included.
All methods used consistent data splits and comparable
inputs within each dataset.
Evaluation metrics included Macro-F1, balanced accuracy,
quadratic weighted kappa, under- and over-triage rates,
triage penalty, expected calibration error, and
risk--coverage metrics.
Results are reported as mean $\pm$ standard deviation
over five random seeds.

\subsection{Overall Performance}

Table~\ref{tab:overall} compares the overall performance
of CRS-Triage with representative baselines on two
independent emergency department datasets at full coverage.
On MIMIC-IV-ED, CRS-Triage achieved the highest Macro-F1
of $0.742$ and QWK of $0.846$, outperforming both
conventional fusion methods and the unified EHR Transformer.
Compared with base evidential fusion, CRS-Triage improved
Macro-F1 from $0.707$ to $0.742$, while reducing the
under-triage rate from $13.1\%$ to $9.6\%$ and the triage
penalty from $0.499$ to $0.406$. It also achieved the
lowest ECE of $0.021$, indicating improved calibration.

A similar trend was observed on MC-MED, where CRS-Triage
achieved a Macro-F1 of $0.718$ and QWK of $0.832$.
Although the overall performance was lower than on
MIMIC-IV-ED, CRS-Triage consistently outperformed
the competing methods. In particular, it reduced the
under-triage rate to $11.2\%$ and the triage penalty
to $0.458$, compared with $14.5\%$ and $0.551$ for
base evidential fusion. These results
suggest that reliability-aware evidence fusion may
improve acuity prediction and reduce asymmetric
triage errors across different clinical settings.

\begin{table*}[!t]
\centering
\begingroup
\fontsize{9}{11}\selectfont
\setlength{\tabcolsep}{2pt}

\begin{tabular}{@{}clcccccc@{}}
\toprule
& & \multicolumn{3}{c}{MIMIC-IV-ED}
& \multicolumn{3}{c}{MC-MED} \\
\cmidrule(lr){3-5}
\cmidrule(lr){6-8}
Coverage & Method
& Error $\downarrow$
& UT $\downarrow$
& Penalty $\downarrow$
& Error $\downarrow$
& UT $\downarrow$
& Penalty $\downarrow$ \\
\midrule

\multirow{5}{*}{80\%}
& Maximum probability
& $18.3\pm0.7$ & $7.0\pm0.6$ & $0.315\pm0.010$
& $20.5\pm0.8$ & $8.2\pm0.6$ & $0.365\pm0.012$ \\

& Evidential certainty
& $16.5\pm0.3$ & $5.9\pm0.4$ & $0.267\pm0.006$
& $18.9\pm0.5$ & $7.0\pm0.4$ & $0.315\pm0.008$ \\

& SelectiveNet
& $16.1\pm0.3$ & $5.7\pm0.3$ & $0.257\pm0.006$
& $18.5\pm0.5$ & $6.8\pm0.4$ & $0.304\pm0.008$ \\

& Cost-aware selection
& $15.8\pm0.4$ & $5.1\pm0.3$ & $0.239\pm0.007$
& $18.2\pm0.5$ & $6.1\pm0.4$ & $0.283\pm0.009$ \\

& \textbf{CRS score}
& $\mathbf{14.7\pm0.3}$
& $\mathbf{4.7\pm0.2}$
& $\mathbf{0.208\pm0.005}$
& $\mathbf{17.1\pm0.4}$
& $\mathbf{5.3\pm0.3}$
& $\mathbf{0.247\pm0.006}$ \\

\midrule

\multirow{5}{*}{90\%}
& Maximum probability
& $22.0\pm0.8$ & $8.4\pm0.7$ & $0.360\pm0.016$
& $24.3\pm0.9$ & $9.7\pm0.7$ & $0.412\pm0.017$ \\

& Evidential certainty
& $20.7\pm0.6$ & $7.7\pm0.5$ & $0.331\pm0.007$
& $23.0\pm0.6$ & $8.9\pm0.5$ & $0.376\pm0.010$ \\

& SelectiveNet
& $20.3\pm0.4$ & $7.4\pm0.4$ & $0.319\pm0.005$
& $22.6\pm0.6$ & $8.6\pm0.5$ & $0.365\pm0.009$ \\

& Cost-aware selection
& $20.0\pm0.5$ & $7.0\pm0.4$ & $0.307\pm0.006$
& $22.3\pm0.6$ & $8.0\pm0.4$ & $0.347\pm0.009$ \\

& \textbf{CRS score}
& $\mathbf{19.1\pm0.4}$
& $\mathbf{6.7\pm0.3}$
& $\mathbf{0.291\pm0.004}$
& $\mathbf{21.4\pm0.5}$
& $\mathbf{7.5\pm0.3}$
& $\mathbf{0.328\pm0.006}$ \\

\bottomrule
\end{tabular}
\endgroup

\caption{Selective prediction on MIMIC-IV-ED and
MC-MED at fixed coverage. Error and UT are in
percent.}
\label{tab:selective}
\end{table*}

\subsection{Selective Prediction and Clinical Risk}


Table~\ref{tab:selective} compares different
selection methods at fixed coverage levels on
MIMIC-IV-ED and MC-MED. Except for SelectiveNet,
all methods use the same predictions generated
by CRS-Triage, allowing their selection strategies
to be compared under a shared predictive model.
The cost-aware selection baseline explicitly
prioritizes cases with lower estimated asymmetric
triage penalties, providing a stronger comparison
than conventional uncertainty-based selection.
On MIMIC-IV-ED, the CRS confidence score achieved
the lowest accepted-case error, under-triage rate,
and triage penalty at both coverage levels.
At $80\%$ coverage, CRS-Triage reduced the
triage penalty from $0.239$ with cost-aware
selection to $0.208$, while decreasing the
under-triage rate from $5.1\%$ to $4.7\%$.
A similar trend was observed at $90\%$ coverage,
where the CRS score achieved a triage penalty
of $0.291$, compared with $0.307$ for
cost-aware selection.

The same pattern was observed on MC-MED.
At $80\%$ coverage, the CRS score achieved
an error rate of $17.1\%$, an under-triage
rate of $5.3\%$, and a triage penalty of
$0.247$, outperforming the competing selection
methods. At $90\%$ coverage, it maintained
the lowest triage penalty of $0.328$.
These results suggest that
combining modality reliability, predictive
uncertainty, and cross-modal consistency
may improve selective triage decisions
across different clinical settings.

\subsection{Ablation and Sensitivity Analysis}

\begin{table}[t]
\centering
\begingroup
\fontsize{9}{11}\selectfont
\setlength{\tabcolsep}{2pt}

\begin{tabular}{@{}lrrr@{}}
\toprule
Variant
& $\Delta$F1
& $\Delta$UT
& $\Delta$P@80 \\
\midrule

\multicolumn{4}{@{}l}{\textit{MIMIC-IV-ED}} \\
w/o reliability
& $-0.018$ & $+1.5$ & $+0.041$ \\
w/o disagreement
& $-0.010$ & $+0.3$ & $+0.023$ \\
Standard selective loss
& $-0.003$ & $+0.8$ & $+0.050$ \\
Equal UT/OT costs
& $+0.002$ & $+3.4$ & $+0.044$ \\

\midrule
\multicolumn{4}{@{}l}{\textit{MC-MED}} \\
w/o reliability
& $-0.019$ & $+1.6$ & $+0.044$ \\
w/o disagreement
& $-0.010$ & $+0.5$ & $+0.022$ \\
\bottomrule
\end{tabular}
\endgroup

\caption{Ablation impact relative to full CRS-Triage.}
\label{tab:ablation}
\end{table}

\begin{table}[t]
\centering
\begingroup
\fontsize{9}{11}\selectfont
\setlength{\tabcolsep}{2pt}

\begin{tabular}{@{}ccccccc@{}}
\toprule
& \multicolumn{3}{c}{MIMIC-IV-ED}
& \multicolumn{3}{c}{MC-MED} \\
\cmidrule(lr){2-4}
\cmidrule(lr){5-7}
Ratio
& F1 $\uparrow$
& UT $\downarrow$
& OT $\downarrow$
& F1 $\uparrow$
& UT $\downarrow$
& OT $\downarrow$ \\
\midrule

$1$
& $0.744$ & $13.0$ & $12.6$
& $0.721$ & $14.7$ & $13.9$ \\

$2$
& $0.743$ & $10.8$ & $14.1$
& $0.720$ & $12.4$ & $15.2$ \\

$3$
& $0.742$ & $9.6$ & $15.1$
& $0.718$ & $11.2$ & $16.3$ \\

$5$
& $0.734$ & $8.0$ & $18.5$
& $0.709$ & $9.4$ & $19.8$ \\

\bottomrule
\end{tabular}
\endgroup

\caption{Sensitivity to the under-/over-triage
penalty ratio.}
\label{tab:cost_sensitivity}
\end{table}

Table~\ref{tab:ablation} evaluates the contribution
of individual components to CRS-Triage. On
MIMIC-IV-ED, removing modality reliability
reduced Macro-F1 from $0.742$ to $0.724$ and
increased the triage penalty from $0.406$ to
$0.451$. The corresponding ECE increased from
$0.021$ to $0.029$, indicating that reliability
modeling contributes to both predictive performance
and calibration. Removing cross-modal disagreement
also increased the triage penalty to $0.433$ at
full coverage and $0.231$ at 80\% coverage,
suggesting that disagreement information improves
the identification of unreliable predictions.

The effect of the selective training objective
was more pronounced at reduced coverage. Replacing
the cost-sensitive objective with standard
classification loss increased the triage penalty
at 80\% coverage from $0.208$ to $0.258$, despite
a relatively small change in full-coverage
Macro-F1. Assigning equal penalties to under-
and over-triage slightly improved Macro-F1 but
increased the under-triage rate from $9.6\%$
to $13.0\%$. This demonstrates the trade-off
between overall classification performance
and asymmetric triage errors.
The core ablations on MC-MED showed similar
patterns. Removing reliability modeling reduced
Macro-F1 from $0.718$ to $0.699$ and increased
the triage penalty from $0.458$ to $0.505$.
Removing disagreement modeling also degraded
performance, suggesting that both components
contribute across different clinical datasets.

Table~\ref{tab:cost_sensitivity} further examines
the effect of the under-/over-triage penalty
ratio. Increasing the penalty on under-triage
consistently reduced under-triage rates, but
increased over-triage and slightly reduced
Macro-F1 at larger ratios. A ratio of $3$
provided a practical empirical trade-off
between predictive performance and asymmetric
triage errors on both datasets.

\subsection{Reliability and Uncertainty Validation}

We evaluated whether the learned reliability,
disagreement, and uncertainty scores reflect
the quality of clinical evidence. Reliability
was assessed using AUROC for detecting incorrect
unimodal predictions, while uncertainty was
evaluated using error-detection AUROC and the
area under the risk--coverage curve (AURC).
To assess sensitivity to degraded evidence,
we randomly masked $50\%$ of observed structured
features or removed $50\%$ of text tokens.
Cross-modal conflict was simulated by replacing
chief complaints with length-matched complaints
from encounters with different acuity labels.
All perturbations were applied at test time
without retraining.

\begin{table}[t]
\centering
\begingroup
\fontsize{9}{11}\selectfont
\setlength{\tabcolsep}{2pt}

\begin{tabular}{@{}lcc@{}}
\toprule
Metric
& MIMIC-IV-ED
& MC-MED \\
\midrule

Structured error AUROC $\uparrow$
& $0.782$ & $0.759$ \\
Text error AUROC $\uparrow$
& $0.805$ & $0.781$ \\
CRS error AUROC $\uparrow$
& $0.812$ & $0.791$ \\
CRS AURC $\downarrow$
& $0.139$ & $0.158$ \\

\midrule

Structured reliability (50\%) $\downarrow$
& $0.684$ & $0.662$ \\
Text reliability (50\%) $\downarrow$
& $0.713$ & $0.691$ \\
Conflict disagreement (50\%) $\uparrow$
& $0.472$ & $0.503$ \\
Conflict detection AUROC $\uparrow$
& $0.823$ & $0.801$ \\

\bottomrule
\end{tabular}
\endgroup

\caption{Reliability and uncertainty validation.}
\label{tab:reliability_validation}
\end{table}

Table~\ref{tab:reliability_validation} shows
that modality reliability scores were informative
about unimodal prediction errors, achieving
AUROC values of $0.782$ and $0.805$ on
MIMIC-IV-ED and $0.759$ and $0.781$ on MC-MED.
The CRS confidence score achieved error-detection
AUROC values of $0.812$ and $0.791$, respectively,
indicating its ability to identify unreliable
fused predictions. Under $50\%$ evidence degradation, normalized
modality reliability decreased substantially
on both datasets. Meanwhile, cross-modal
disagreement increased under conflicting
chief complaints, with conflict-detection
AUROC values of $0.823$ and $0.801$.
These results suggest that the proposed
mechanisms capture complementary aspects
of incomplete and inconsistent clinical
evidence, supporting more reliable
selective triage.

\section{Conclusion}
In this work, we proposed CRS-Triage, a confidence-
and reliability-aware selective framework for
emergency triage under incomplete and unreliable
clinical evidence. By integrating modality reliability,
cross-modal disagreement, and predictive uncertainty,
CRS-Triage improves multimodal acuity prediction
and supports selective deferral of uncertain cases.
An asymmetric cost-sensitive objective further
reduces clinically undesirable under-triage errors.
Experiments on MIMIC-IV-ED and MC-MED demonstrate
improved predictive performance, reduced triage
penalties, and more effective selective prediction.
Additional analyses validate the ability of the
proposed reliability and uncertainty mechanisms
to identify unreliable and conflicting evidence.
Future work will investigate adaptive reliability
estimation under changing clinical data distributions.

\section{
Acknowledgement} Guan Qiang received trainee support through a Mitacs Accelerate Project involving Kelello Health, the Centre for Addiction and Mental Health (CAMH), and Western University (Project ID GMS-100025).

\bibliography{aaai2026}

\end{document}